\documentclass[10pt,twocolumn,letterpaper]{article}
 
\usepackage{iccv}
\usepackage{times}
\usepackage{epsfig}
\usepackage{graphicx}
\usepackage{amsmath}
\usepackage{amssymb}

\usepackage{multirow}
\usepackage{amsfonts}
\usepackage{color}
\usepackage{subfigure}
\usepackage{adjustbox,lipsum}
 
\usepackage{dsfont}
\usepackage{booktabs}
\usepackage{authblk}
 
\usepackage[pagebackref=true,breaklinks=true,letterpaper=true,colorlinks,bookmarks=false]{hyperref}
 
\iccvfinalcopy 
 
\def\iccvPaperID{****} 

\begin{document}
 
\title{A Novel Unsupervised Camera-aware Domain Adaptation Framework for Person Re-identification}
 
\author[1]{Lei Qi}
\author[2$^*$]{Lei Wang}
\author[1]{Jing Huo}
\author[3]{Luping Zhou}
\author[1]{Yinghuan Shi}
\author[1\thanks{Corresponding authors: Yang Gao; Lei Wang. The work of J. Huo was supported by NSFC (61806092) and Jiangsu Natural Science Foundation (BK20180326). The work of Y. Shi was supported by the Fundamental Research Funds for the Central Universities (020214380056), NSFC (61673203), CCF-Tencent Open Research Fund (RAGR20180114). The work of Y. Gao was supported by NSFC (61432008).}]{Yang Gao}
 
\affil[1]{State Key Laboratory for Novel Software Technology, Nanjing University}
\affil[2]{School of Computing and Information Technology, University of Wollongong}
\affil[3]{School of Electrical and Information Engineering, The University of Sydney}

\maketitle

\begin{abstract}
Unsupervised cross-domain person re-identification (Re-ID) faces two key issues. One is the data distribution discrepancy between source and target domains, and the other is the lack of label information in target domain. They are addressed in this paper from the perspective of representation learning. 
For the first issue, we highlight the presence of camera-level sub-domains as a unique characteristic of person Re-ID, and develop ``camera-aware'' domain adaptation to reduce the discrepancy not only between source and target domains but also across these sub-domains.
For the second issue, we exploit the temporal continuity in each camera of target domain to create discriminative information. This is implemented by dynamically generating online triplets within each batch, in order to maximally take advantage of the steadily improved feature representation in training process. Together, the above two methods give rise to a novel unsupervised deep domain adaptation framework for person Re-ID. Experiments and ablation studies on benchmark datasets demonstrate its superiority and interesting properties.
\end{abstract}
 
\vspace{-15pt}
\section{Introduction}
Person re-identification (Re-ID) matches images of the same identity captured by different cameras of non-overlapping views~\cite{DBLP:journals/tmm/ChenLLCH11}. 
In unsupervised cross-domain person Re-ID, labeled data is only available in source domain, while all data in target domain is unlabeled~\cite{qi2018unsupervised}. 
It aims to learn an effective model to conduct Re-ID in target domain.
 
Unsupervised cross-domain person Re-ID faces two key issues. One is the data distribution discrepancy of source and target domains, caused by the variations such as body pose, view angle, illumination, image resolution, occlusion and background. The other is the lack of label information in target domain, due to time-consuming or even infeasible manual annotation in real applications. 
Tangible progress has been made to address them~\cite{DBLP:journals/tcsv/WangZLZ16,qi2018unsupervised,wei2018person,deng2018image,zhong2018generalizing,Bak_2018_ECCV}, as reviewed in next section. However, unsupervised cross-domain person Re-ID is still far from satisfactory, especially when compared with the supervised counterpart~\cite{DBLP:conf/cvpr/AhmedJM15,DBLP:conf/iccv/ZhengZY17,kalayeh2018human}. 
 
This paper aims to better address the above two issues from the perspective of representation learning. We try to learn a deep feature representation (or equally, a shared subspace) in which the discrepancy of source and target domains is more effectively reduced and the discriminative information in target domain is more effectively reflected. 
 
First, in most practical applications, the camera ID of a frame can be readily obtained\footnote{For example, camera ID is provided for each frame in benchmark datasets of person Re-ID, as will be shown in the experiment.}. The frames from the same camera exhibit a common visual style (e.g., in terms of background, image resolution, viewing angle, and so on). This camera-specific style has recently started attracting attention in person Re-ID (e.g., image-image translation)～\cite{zhong2018generalizing}. 
The presence of camera-level \textit{sub-domains} is a unique characteristic of cross-domain person Re-ID. Nevertheless, it has not been well exploited from the perspective of representation learning. 
To utilize this characteristic, we propose a novel \textit{camera-aware} domain adaptation. It emphasizes that with the learned representation, the distribution discrepancy across camera-level sub-domains shall also be sufficiently reduced, dealing with the discrepancy of source and target domains at a finer level. 
This idea is realized by adversarial learning with a novel criterion called \textit{cross-domain camera equiprobability}. 
 
 
 
Second, temporal information is usually available for the images from a camera (e.g., via frame ID) in person Re-ID. 
Temporally close frames more likely correspond to the same subject, 
which has long been used as an important cue for video analysis. Nevertheless, it has not been well exploited for unsupervised cross-domain person Re-ID to handle the lack of label information in target domain. In this paper, temporal information is jointly used with image distances to generate, in an unsupervised manner, triplets from each camera in target domain. A smart scheme is proposed to better seek true positive and true negative images. More importantly, instead of generating triplets offline, we dynamically generate triplets online in each batch during training. This allows us to fully take advantage of the steadily improved feature representation to produce better triplets. In turn, these triplets help networks to learn better feature representations, forming a positive loop to exploit discriminative information from target domain. 
 
Together, the above two improvements give rise to a novel domain adaptation framework for unsupervised cross-domain person Re-ID. As will be shown, both improvements are essential and it is their joint effort that makes the proposed framework really function. For clarity, the contributions in this work are summarized as follows. 
 
First, a camera-aware domain adaptation method is developed by considering the unique presence of camera-level sub-domains in person Re-ID. 
To the best of our survey, our adversarial learning based method is the first one, among those aiming to learn better feature representation, to integrate source and target domains at this fine level.
 
Second, an unsupervised online in-batch triplet generation method is proposed to explore the underlying discriminative information in unlabeled target domain. 
Through high-quality triplets, it provides important information to boost the performance of the whole framework.  
 
Last, both theoretical analysis and experimental studies are conducted to illustrate the proposed camera-aware domain adaptation. The results and ablation studies demonstrate the superiority of the proposed framework and its interesting properties.
\section{Related Work}
 
\textbf{Unsupervised domain adaptation.} Unsupervised cross-domain person Re-ID is related to unsupervised domain adaptation, a more general technique handling unlabeled target domain with the help of labeled source domain. 
In the literature, most unsupervised domain adaptation methods learn a common mapping between source and target distributions. 
Several methods based on the maximum mean discrepancy (MMD) have been proposed~\cite{DBLP:conf/icml/LongC0J15,NIPS2016_6110,DBLP:journals/corr/ZhangYCW15,DBLP:journals/corr/TzengHZSD14}.
Long \textit{et al.}~\cite{DBLP:conf/icml/LongC0J15} use a new deep adaptation network, where hidden representations of all task-specific layers are embedded in a Reproducing Kernel Hilbert space. To transfer a classifier from source domain to target domain, the work in~\cite{NIPS2016_6110} jointly learns adaptive classifiers between the two domains by a residual function. In~\cite{DBLP:conf/eccv/GhifaryKZBL16,DBLP:conf/nips/BousmalisTSKE16}, autoencode-based methods are investigated to explore the discriminative information in target domain. Recently, adversarial learning~\cite{DBLP:conf/icml/GaninL15,DBLP:journals/corr/abs-1803-09210,DBLP:conf/cvpr/TzengHSD17} has been applied to domain adaptation. Ganin \textit{et al.}~\cite{DBLP:conf/icml/GaninL15} propose the gradient reversal layer (GRL) to pursue the same distribution between source and target domains. Inspired by generative adversarial networks (GANs),
Tzeng \textit{et al.}~\cite{DBLP:conf/cvpr/TzengHSD17} leverage a GAN loss to match the data distributions of source and target domains. 
 
Nevertheless, for a person Re-ID task, the distribution discrepancy also exists at the camera level. It will not be effectively reduced when only the overall domain-level discrepancy is concerned. 
In this sense, directly applying existing unsupervised domain adaptation methods to a person Re-ID task may not be the best option.
 
\textbf{Unsupervised cross-domain person Re-ID.} As previously mentioned, most existing methods on this topic address two issues, \textit{i.e.,} reducing data distribution discrepancy between two domains and generating discriminative information for target domain. In the literature, methods have been developed to learn a shared subspace or dictionary across domains~\cite{DBLP:conf/cvpr/PengXWPGHT16, DBLP:journals/tcsv/WangZLZ16, DBLP:journals/tip/MaLYL15,qi2018unsupervised}. However, these methods are not based on deep learning and thus do not fully explore the high-level semantics in images. Recently, several deep-learning-based methods~\cite{zhong2018generalizing,Bak_2018_ECCV,wang2018transferable,lv2018unsupervised} have been seen. In~\cite{lv2018unsupervised,wang2018transferable}, generating pseudo labels for target images is investigated. Lv \textit{et al.}~\cite{lv2018unsupervised} propose an unsupervised incremental learning algorithm, aided by the transfer learning of spatio-temporal patterns of pedestrians in target domain. In~\cite{wang2018transferable}, the proposed approach simultaneously learns an attribute-semantic and identity-discriminative feature representation in target domain. However, when generating discriminative information for target domain, the above methods do not utilize temporal continuity of images in each camera. Moreover, the generation of information is usually conducted offline and separately, instead of on the fly during training. All of these will be improved in our framework. 
 
 
Recently, 
generating extra training images for target domain has become popular~\cite{wei2018person,deng2018image,zhong2018generalizing,Bak_2018_ECCV}. 
Wei \textit{et al.}~\cite{wei2018person} impose constraints to maintain the identity in image generation. The approach in~\cite{deng2018image} enforces the self-similarity of an image before and after translation and the domain-dissimilarity of a translated source image and a target image. Zhong \textit{et al.}~\cite{zhong2018generalizing} propose to seek camera invariance by using unlabeled target images and their camera-style transferred counterparts as positive matching pairs. Besides, it views source and target images as negative pairs for the domain connectedness. Note that these methods have attempted to deal with the distribution discrepancy at the camera level. Differently, they reduce the discrepancy through the approach of image generation, rather than learning better representation as in this paper. As will be demonstrated in the experiment, our approach can produce better person Re-ID performance in target domain than these methods.
 
 
\section{The Proposed Framework}
 
Our framework consists of three objectives, including i) classification of the labeled images in source domain; ii) camera-aware domain adaptation via adversarial learning; and iii) enforcing discrimination information in target domain. The first objective is not our focus and is implemented by following the literature~\cite{DBLP:conf/cvpr/TzengHSD17,DBLP:conf/icml/LongC0J15}. The second and third objectives are detailed in Sections~\ref{sec:CAL} and~\ref{sec:UOT}. 

\subsection{Camera-aware domain adaptation}\label{sec:CAL}
 
In person Re-ID, images of different cameras form a set of unique units. This is also reflected in cross-domain discrepancy. 
Merely reducing the overall discrepancy of source and target domains will not effectively handle the camera-level discrepancy, and this could adversely affect the quality of learned feature representation. 
We put forward a camera-aware domain adaptation to map the images of different cameras into a shared subspace. To achieve this, a camera-aware adversarial learning (CAL in short) method is developed. Unlike conventional adversarial learning dealing with two domains~\cite{DBLP:conf/icml/GaninL15}, CAL deals with multiple sub-domains (i.e., cameras in source and target domains). 
 
Let $X_{s}$ and $X_{t}$ be the training images in source and target domains and $X=[X_{s}, X_{t}]$. The camera IDs (i.e., the label of each camera class) of the images in $X$ are denoted by $Y_{c}$. Let $C_{s}$ and $C_{t}$ denote the number of the cameras in source and target domains, respectively, and $C = C_s + C_t$. Adversarial learning involves the optimization of discriminator and generator. As commonly seen, the discriminator in this work is optimized by a cross-entropy loss defined on the $C$ camera classes in source and target domains as
\begin{equation}\label{eq11}
\begin{aligned}
&\min_{D}\mathcal{L}_\mathrm{CAL-D}(X, Y_{c}, B)= \\
&\min_{D}\left[\mathbb{E}_{(x,y_{c})\sim (X,Y_{c})}\left(-\sum_{k=1}^{C}\delta(y_{c}-k)\log D(B(x), k)\right)\right],
 \end{aligned}
\end{equation} where $x$ denotes an image, $y_{c}$ the camera class label of $x$, and $\delta(\cdot)$ the Dirac delta function. $B$ denotes the backbone network, and $B(x)$ is the feature representation of $x$. $D$ denotes the discriminator and $D(B(x), k)$ denotes the prediction score for $x$ with respect to the $k$-th camera class.
 
The generator in this work is the backbone network $B$. Due to the special need to deal with multiple camera classes, the optimization of $B$ becomes tricky. We first investigate the gradient reversal layer (GRL) technique~\cite{DBLP:conf/icml/GaninL15}. By showing its limitations for this task, we propose a new criterion ``cross-domain camera equiprobability'' (CCE) for our task. 
 
\textbf{GRL-based adaptation scheme.} The gradient reversal layer (GRL)~\cite{DBLP:conf/icml/GaninL15} is commonly used to reduce distribution discrepancy of two domains by \textit{maximizing} the domain discrimination loss (i.e., Eq.(\ref{eq11})). A direct extension of GRL to our task leads to optimizing the generator $B$ as
\begin{equation}\label{eq12}
\begin{aligned}
&\min_{B}\mathcal{L}_\mathrm{CAL-B}(X, Y_{c}, D)\triangleq\max_{B}\mathcal{L}_\mathrm{CAL-D}(X, Y_{c}, D) \\
&=\min_{B}\left[\mathbb{E}_{(x,y_{c})\sim (X,Y_{c})}\sum_{k=1}^{C}\delta(y_{c}-k)\log D(B(x), k)\right],
 \end{aligned}
\end{equation}where for consistency it is written as minimizing the negative discriminator loss. 
 
To train the backbone network $B$ with Eq.(\ref{eq12}), we insert GRL between $B$ and $D$ as in the literature~\cite{DBLP:conf/icml/GaninL15}. During forward propagation, GRL is simply an identity transform. During backpropagation, GRL reverses (i.e., multiplying by a negative constant) the gradients of the domain discriminator loss with respect to the network parameters in feature extraction layers and pass them backwards. This GRL-based adaptation scheme can somehow work to reduce distribution discrepancy across different cameras (i.e., sub-domains), as will be experimentally demonstrated shortly. 
 
However, this scheme has a drawback. Maximizing the discriminator loss only enforces an image not to be classified into its true camera class. It will appear to be ``equivalently good'' for this optimization as long as an image is classified into any wrong camera classes, including those from the \textit{same} domain. 
In this case, this scheme will not be able to effectively pull source and target domains together. The larger the discrepancy between source and target domains is, the more pronounced this issue could be. 
 
\begin{figure}
\centering
\includegraphics[width=8cm]{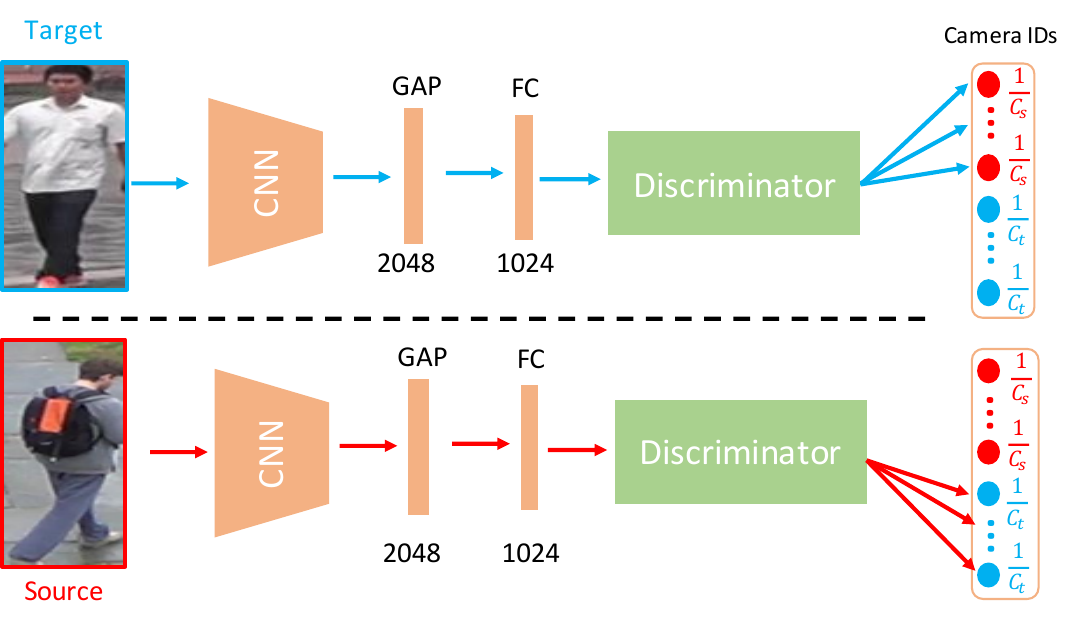}
\caption{Illustration of training the backbone network $B$ with the proposed cross-domain camera equiprobability (CCE) loss in our CAL method at each iteration. FC and GAP stand for fully connected layer and global average pooling. The top of the figure shows that each image in target domain (in blue) is required to be equiprobably misclassified into all camera classes in source domain (in red). The bottom similarly shows the case for each image in source domain. The discriminator is set as a 128-d FC layer.}​
\label{fig3}
\vspace*{-20pt}
\end{figure}

\textbf{CCE-based adaptation scheme.} 
In this scheme, we enforce ``preferred misclassification patterns'' to maximize the discriminator loss. Noting that the primary goal of cross-domain person Re-ID is to reduce the distribution discrepancy of source and target domains, we require that with the learned feature representation, an image from a camera class in source domain shall be wrongly classified into a camera class in target domain and vice versa. 
As we do not have any bias on the camera classes in the opposite domain, it is required that the image shall be misclassified into them with equal probability, as illustrated in Fig.~\ref{fig3}. This is where the name ``cross-domain camera equiprobability'' (CCE) comes from. This scheme effectively avoids considering the misclassification among the camera classes of the same domain, which is unfavourably done in the GRL scheme, and makes specific efforts to pull source and target domains together.
Also, the CCE-based scheme can practically lead to alleviating the discrepancy across all cameras in target domain, as demonstrated in the experiment.
 
 
Let ${\mathcal S}$ and ${\mathcal T}$ denote source and target domains. Formally, the CCE loss on an image $x$ can be expressed as
\begin{equation}\label{eq13}
\begin{aligned}
\mathcal{L}_\mathrm{CCE}(x)=\left\{\begin{matrix}
 -\frac{1}{C_{s}}\sum_{i=1}^{C_{s}}\log (D(B(x),i)),~x \in {\mathcal T} \\ 
 -\frac{1}{C_{t}}\sum_{j=1}^{C_{t}}\log (D(B(x),j)),~x \in {\mathcal S}
\end{matrix}\right.
\end{aligned}
\end{equation}
For $x$ in target domain, $D(B(x),i)$ denotes the predicted probability that $x$ belongs to the $i$th camera class in source domain. Similar definition applies to $D(B(x),j)$ for $x$ in source domain. In this way, the optimization for training the backbone network $B$ (as a generator) is defined as
\begin{equation}\label{eq14}
\begin{aligned}
\min_{B}\mathcal{L}_\mathrm{CAL-B}(X, D)=\min_{B}\mathbb{E}_{x\sim X} \mathcal{L}_\mathrm{CCE}(x, D).
 \end{aligned}
\end{equation} 
 
During adversarial training, $D$ and $B$ are trained in an alternate way. Each iteration consists of two steps: i) Temporarily fixing the weights of $B$, $D$ is trained by Eq.(\ref{eq11}) to predict the camera ID of each image; ii) Temporarily fixing the weights of $D$, $B$ is trained by Eq.(\ref{eq14}) to learn feature representation. This is repeated till convergence. Note that traditional two-domain adversarial learning~\cite{DBLP:conf/icml/GaninL15,DBLP:conf/cvpr/TzengHSD17} is just a special case of this CCE-based scheme when there is exactly one camera class in each of source and target domains.
 
\textit{Remarks.} 
A question may arise: why do we prefer the CCE criterion rather than simply requiring an image from each camera class to be equiprobably misclassified into all the other (i.e., $C_{s}+C_{t}-1$) camera classes? This is because in person Re-ID the cross-domain discrepancy is usually more significant than the within-domain counterpart and affects the performance more. By enforcing the cross-domain camera equiprobability, the CCE criterion puts higher priority on reducing the former discrepancy and therefore is a better option. Its effectiveness and advantage will be validated in the experimental study.     
 
\textbf{Theoretical analysis of CCE.} At last, we provide theoretical analysis to gain more insight on this new criterion. The full analysis is provided in the supplement material. 
\textbf{Proposition.} \textit{Let ${\mathcal S}$ and ${\mathcal T}$ denote source and target domains. $x^s$ and $x^t$ are the images from the two domains; $p_s(x)$ and $p_t(x)$ are their probability density functions; and $C_s$ and $C_t$ are the number of camera classes in these two domains. Let $p(x|\mathcal{C}^s_i)$ and $p(x|\mathcal{C}^t_i)$ be the class-conditional density functions of the $i$th camera class in source and target domains, respectively. It can be proved (see supplement material) that ideally, minimizing the CCE loss will lead to}
\begin{eqnarray}\label{eqn:proposition} 
p(x^s|\mathcal{C}^t_i) &=& p_s(x^s),\quad \forall x^s \in {\mathcal S};~i=1,\cdots,C_t.\\~\nonumber
p(x^t|\mathcal{C}^s_i) &=& p_t(x^t),\quad \forall x^t \in {\mathcal T};~i=1,\cdots,C_s.\\~\nonumber
p_s(x) &=& p_t(x), \quad \forall x \in {\mathcal S}\cup{\mathcal T}.
\end{eqnarray}This result indicates that in the learned shared space: 
 
i) For any image in source domain, it will not feel the discrepancy among the $C_t$ camera classes in target domain. Its class-conditional density function value for these camera classes (e.g., $p(x^s|\mathcal{C}^t_i)$) just equals its density function value in its own domain (e.g., $p_s(x^s)$). The above conclusion also applies to any image in target domain in a similar but reverse way. 
 
 
ii) The data distributions of source and target domains, $p_s(x)$ and $p_t(x)$, will become identical and the overall cross-domain distribution discrepancy can be removed. 
 
Meanwhile, it is worth mentioning that minimizing the CCE loss does not theoretically guarantee that an image (in either source or target domain) will not feel the distribution discrepancy among the camera classes in \textit{its own} domain. 
Nevertheless, jointly considering the above three proved equalities, it can be reasonably expected (see the supplement) that the above situation could be observed in practice. This will be experimentally demonstrated shortly.  
 

\subsection{Unsupervised online triplet generation}~\label{sec:UOT}
Only reducing cross-domain distribution discrepancy is insufficient, even though the above camera-aware domain adaptation is deployed. Rather, maintaining the intrinsic properties of target domain is equally essential. Otherwise, the distribution of target domain could be arbitrarily altered just for reducing the distribution discrepancy, significantly degrading the Re-ID performance in target domain. To avoid this, this framework explores the underlying discriminative information in target domain. 
 
This information is explored in the form of image triplet, consisting of an anchor image, a positive image (i.e., sharing the same identity as the anchor) and a negative image (i.e., having a different identity). When selecting positive and negative images, we not only consider the distance between images but also jointly utilize the temporal information among images, which can often be obtained via the frame ID in person Re-ID.
More importantly, we generate triplets online in each batch during training. This allows triplet generation to effectively take advantage of the steadily improved feature representation to produce better triplets.
Note that the triplet generation is carried out in an unsupervised manner and only needed in training process. 
 
Given a camera in target domain, all of its images are sorted temporally into a list. From this list, $p$ non-overlapping fragments are randomly selected to construct a batch. Each fragment consists of $q$ images, and the batch therefore contains $n$ ($=p\times q$) images in total. Each of the $n$ images is used as an anchor image to create triplets in turn. 
To begin with, a pairwise distance matrix ${\mathbf M}\in \mathbb{R}^{n\times n}$ is computed for the $n$ images, with the feature representation learned by the network so far. To generate triplets for an anchor image $I_{a}$, we develop the following rules.
 
Above all, according to ${\mathbf M}$, sort all the $(n-1)$ images (excluding $I_{a}$) in the batch in ascending order of the distance from $I_{a}$. The obtained list is denoted by $S(I_{a})$.
 
\textbf{Positive image selection.} To be selected as positive, an image must meet both the requirements: i) it is within the top-$k$ positions of $S(I_{a})$, {and} ii) it is from the same fragment as $I_{a}$. The first requirement ensures that this image is indeed similar to $I_{a}$ in terms of feature representation, while the second one further increases its likelihood of positiveness with temporal information. Jointly using these two requirements helps us to select highly likely (not guaranteed though) true positive images. In implementation, $k$ is empirically set. The total number of selected positive images is denoted by $k_p$. Note that $k_p$ could be zero, meaning that this anchor cannot find any positive images by the above rule. In this case, this anchor will not be taken into account.


\textbf{Negative image selection.} 
Starting from the head of the list $S(I_{a})$, each image $I$ is checked in turn against the following conditions: 1) $I$ is \textit{not} from the same fragment of $I_{a}$, and 2) no image in the fragment of $I$ has previously been selected as negative. That is, the negative images are selected as the nearest neighbours of the anchor from the fragments \textit{other than} the anchor's, with the \textit{condition} that each of these negative images shall be, respectively, collected from a \textit{different} fragment. Such a rule is designed to deal with the case that the same person may reappear in two or more fragments. Requiring each negative sample to reside in different fragments well reduces (although cannot entirely avoid) the chance to mis-select a truly positive sample as negative. The total number of selected negative images is denoted by $k_n$. 
 
Once triplets are generated in a training batch, we can train the backbone network via a triplet loss defined as
\begin{equation}\label{eq02}
\begin{aligned}
\mathcal{L}_{\mathrm{Triplet}}=\sum_{a=1}^{n} w_{a}[\bar{d_{p}}(I_{a})-\bar{d_{n}}(I_{a}) +m]_{+},
 \end{aligned}
 \vspace*{-1pt}
\end{equation}where $w_a$ is zero if $I_a$ has no positive images and one otherwise.
$[t]_{+}$ equals $t$ if $t > 0$ and zero otherwise. $m$ is the margin. $\bar{d_{p}}(I_{a})=\frac{1}{k_{p}} \sum_{i=1}^{k_{p}}d(I_{a},I_{p}^{i})$ and $\bar{d_{n}}(I_{a})=\frac{1}{k_{n}} \sum_{i=1}^{k_{n}}d(I_{a},I_{n}^{i})$ are the average distances of the positive and negative samples from the anchor, respectively. Using average distances here helps to mitigate the adverse effect in case any positive or negative sample is wrongly selected. $d(\cdot,\cdot)$ is just the distance used for the distance matrix ${\mathbf M}$.
 
Finally, although ${\mathbf M}$ can be simply calculated by Euclidean distance, more advanced measures can be readily used. This work uses a re-ranking algorithm~\cite{DBLP:conf/cvpr/ZhongZCL17} to improve ${\mathbf M}$, in order to generate even better triplets and further improve the person Re-ID performance in target domain.

\subsection{The overall proposed framework}\label{sec:TOPF}
\begin{figure}
\centering
\includegraphics[width=8cm]{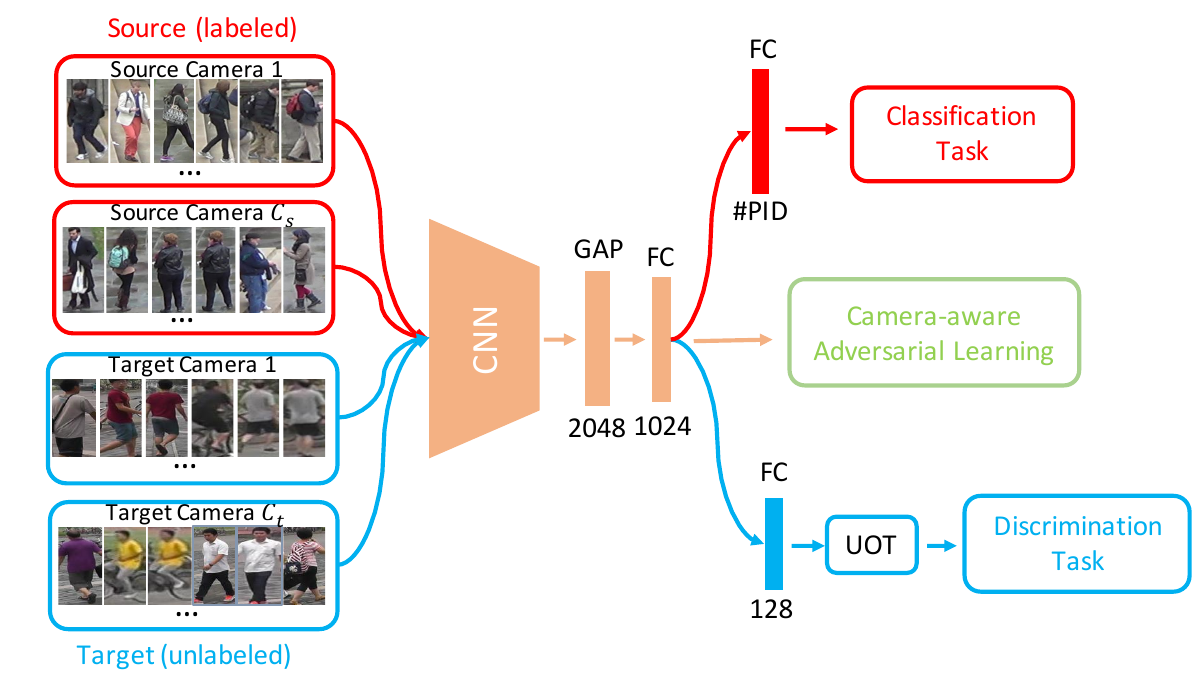}
\caption{Illustration of the proposed unsupervised camera-aware domain adaptation framework, where FC and GAP denote fully connected layer and global average pooling. \#PID and UOT denote the total number of person classes in source domain and unsupervised online triplet generation, respectively.}
\label{fig1}
\vspace*{-15pt}%
\end{figure}
 
Recall that this framework consists of an adversarial task across the cameras (sub-domains) in source and target domains, a discrimination task for target domain, and a classification task for source domain, as shown in Fig.~\ref{fig1}. 
The overall loss function in a training batch is expressed as
\begin{equation}\label{eq09}
\begin{aligned}
&\min_{D}\mathcal{L}_\mathrm{CAL-D}(X, Y_c, B)= \\
&\min_{D}\left[\mathbb{E}_{(x,y_c)\sim (X,Y_C)}\left(-\sum_{k=1}^{C}\delta(y_c-k)\log D(B(x), k)\right)\right],\\
&\min_{B}\mathcal{L}(X,Z_{s},D)=\\
&\min_{B}\left(\mathcal{L}_\mathrm{Cross}(X_{s},Z_{s})+\lambda_{1} \mathcal{L}_\mathrm{Triplet}(X_{t})+\lambda_{2}\mathcal{L}_\mathrm{CAL-B}(X,D)\right),
\end{aligned}
\end{equation}where $\mathcal{L}_\mathrm{Cross}$, $\mathcal{L}_\mathrm{Triplet}$ and $\mathcal{L}_\mathrm{CAL-B}$ are the cross-entropy loss for source domain, the triplet loss for target domain, and the adversarial loss for $B$. $\lambda_{1}$ and $\lambda_{2}$ are the trade-off parameters. $Z_{s}$ is the person IDs of $X_{s}$ in source domain. To caclulate $\mathcal{L}_\mathrm{Triplet}$, one camera in target domain is randomly chosen to construct the training batch and generate triplets at each iteration.
 
In this framework, ResNet-50~\cite{DBLP:conf/cvpr/HeZRS16} is used as backbone network. Global average pooling (GAP) is used to obtain the 2048-d feature representation. 
To do person Re-ID in target domain, the 2048-d feature representation is extracted for each query and gallery images, and an $L_{2}$ normalization is applied. Euclidean distance is calculated to rank the gallery images for a query. 

\section{Experiments}
\subsection{Datasets and settings}\label{sec:EXP-DS}
We evaluate our approach on two large-scale benchmark datasets: Market1501~\cite{DBLP:conf/iccv/ZhengSTWWT15}, DukeMTMC-reID (Duke in short)~\cite{DBLP:conf/iccv/ZhengZY17}.
 \textbf{Market1501} contains 1,501 persons with 32,668 images from six cameras. Among them, 12,936 images of 751 identities are used for training. For evaluation, there are 3,368 and 19,732 images in the query and gallery sets. \textbf{DukeMTMC-reID} has 1,404 persons from eight cameras, with 16,522 training images of 702 identities, 2,228 queries, and 17,661 gallery images. Both camera ID and Frame ID information are available on Market1501 and DukeMTMC-reID. To evaluate person Re-ID performance, we use Rank-1 accuracy and mAP~\cite{DBLP:conf/iccv/ZhengSTWWT15}. On Market1501, there are single- and multi-query evaluation protocols. We use the more challenging single-query protocol.
 
For training CAL, we randomly select 64 images from each of source and target domains in a batch. The 64 images from source domain are also used to train the classification component. To generate triplets, we set $p$ (i.e., number of temporal fragments) and $q$ (i.e., number of images per fragment) to 12 and 10. $k$ and $k_{n}$ used in the selection of positive and negative samples are set as $5$ and $2$. The margin of triplet loss, $m$, is 0.3. $\lambda_{1}$ and $\lambda_{2}$ in Eq.(\ref{eq09}) are set as 1. 
 
The proposed model is trained with the SGD optimizer in a total of 100 epochs. The initial learning rates of the fine-tuned parameters (those in the pre-trained ResNet-50 on ImageNet~\cite{DBLP:conf/cvpr/DengDSLL009}) and the new parameters (those in the newly added layers) are 0.1 and 0.01, respectively. When the number of epochs exceeds 80, we decrease the learning rates by a factor of 0.1. The size of the input images is $256 \times 128$. 
 
Note that the \textbf{baseline} model (denoted by BL) in this experiment represents the ResNet-50~\cite{DBLP:conf/cvpr/HeZRS16} with additional 1024-d fully connected (FC) layer and cross-entropy loss. UOT and UOT(eud) denote two unsupervised online triplet generation variants in which the aforementioned re-ranking algorithm~\cite{DBLP:conf/cvpr/ZhongZCL17} (our default setting) and Euclidean distance are used to compute the distance matrix ${\mathbf M}$, respectively. 
CAL-GRL and CAL-CCE denote the proposed camera-aware adversarial learning (CAL) implemented by the GRL and CCE schemes, respectively. For clarity, we use UCDA-GRL\footnote{UCDA is short for Unsupervised Camera-aware Domain Adaptation.} and UCDA-CCE to represent BL+UOT+CAL-GRL and BL+UOT+CAL-CCE, respectively. Note that for the proposed framework, camera IDs and frame IDs are only needed in training, but not in the test stage.

\subsection{Comparison with the state-of-the-art methods}\label{sec:EXP-CSM}
We compare our approach with seven state-of-the-art unsupervised cross-domain person Re-ID approaches. 
Among them, there are two non-deep-learning-based methods (UMDL~\cite{DBLP:conf/cvpr/PengXWPGHT16} and UJSDL~\cite{qi2018unsupervised}) and five deep-learning-based methods. The latter includes two recent pseudo-label-generation-based methods (TFusion~\cite{lv2018unsupervised} and TJ-AIDL~\cite{wang2018transferable}) and three recent image-generation-based approaches (PTGAN~\cite{wei2018person}, SPGAN~\cite{deng2018image} and HHL~\cite{zhong2018generalizing}). Market1501 and Duke are in turn used as source and target domains to compare these methods. Table~\ref{tab04} reports the result. As seen, our approach (i.e., UCDA-CCE) consistently achieves the best results in both settings. For the setting of ``Duke$\rightarrow$Market1501'', it obtains $34.5\%$ and $64.3\%$ in mAP and Rank-1 accuracy, outperforming all the methods in comparison. In the setting of ``Market1501$\rightarrow$Duke'', UCDA-CCE still excels. Particularly, compared with HHL~\cite{zhong2018generalizing}, the state-of-the-art by using image generation to reduce the camera-level discrepancy in target domain, UCDA-CCE gains $9.5\%$ ($36.7$ vs. $27.2$) in mAP and $8.5\%$ (55.4 vs. 46.9) in Rank-1 accuracy. 
 
This result shows the advantage of our approach over the pseudo-label-generation-based and image-generation-based methods. It effectively alleviates the data distribution discrepancy at the camera level through representation learning. 
We will demonstrate this property in Section~\ref{sec:EXP-FECAL}.
 
\renewcommand{\cmidrulesep}{0mm} 
\setlength{\aboverulesep}{0mm} 
\setlength{\belowrulesep}{0mm} 
\setlength{\abovetopsep}{0cm} 
\setlength{\belowbottomsep}{0cm}
\begin{table}[htbp]
 \centering
 \caption{Comparison with the state-of-the-art methods of unsupervised Re-ID on Market1501 and DukeMTMC-reID (Duke). }
 \begin{adjustbox}{width=0.45\textwidth}
   \begin{tabular}{|c|cc|cc|}
   \toprule
   \multirow{2}[2]{*}{Method} & \multicolumn{2}{c|}{Duke$\rightarrow$Market1501} & \multicolumn{2}{c|}{Market1501$\rightarrow$Duke} \\
\cmidrule{2-5}         & mAP  & Rank-1 & mAP  & Rank-1 \\
   \midrule
   UMDL~\cite{DBLP:conf/cvpr/PengXWPGHT16} & 12.4 & 34.5 & 7.3  & 18.5 \\
   UJSDL~\cite{qi2018unsupervised} & - & 50.9 & -     & 32.2 \\
   \midrule
   TFusion~\cite{lv2018unsupervised}  & - & 60.8 & -     & - \\
   TJ-AIDL~\cite{wang2018transferable} & 26.5 & 58.2 & 23   & 44.3 \\
   \midrule
   PTGAN~\cite{wei2018person} & -     & 38.6 & -     & 27.2 \\
   SPGAN+LMP~\cite{deng2018image} & 26.7 & 57.7 & 26.2 & 46.4 \\
   HHL~\cite{zhong2018generalizing}  & 31.4 & 62.2 & 27.2 & 46.9 \\
   \midrule
    UCDA-GRL & 30.9 & 60.4 & 31.0 & 47.7 \\
   UCDA-CCE (Ours) & \textcolor[rgb]{ 1, 0, 0}{\textbf{34.5}} & \textcolor[rgb]{ 1, 0, 0}{\textbf{64.3}} & \textcolor[rgb]{ 1, 0, 0}{\textbf{36.7}} & \textcolor[rgb]{ 1, 0, 0}{\textbf{55.4}} \\
   \bottomrule
   \end{tabular}%
   \end{adjustbox}
 \label{tab04}%
 \vspace*{-5pt}
\end{table}%
 
Our approach can be easily extended to semi-supervised person Re-ID. That is, when within each camera in target domain, the labels of person identity become available for each frame (e.g., obtained by tracking algorithms or manual annotation), our approach can utilize this information by simply changing our unsupervised online triplet generation (UOT) to traditional triplet generation~\cite{hermans2017defense}. We name this setting SOT where ``S'' stands for semi-supervised.\footnote{This setting is called semi-supervised because person identity labels are only available \textit{within} each individual camera but not across all cameras.} Our approach is compared with TAUDL~\cite{Li_2018_ECCV}, a recent state-of-the-art method for video-based person Re-ID. Note that TAUDL utilizes the person identity labels within each camera in target domain but does not employ cross-domain adaptation from a source domain.
This comparison aims to show that via CAL, our approach can effectively utilize source domain to produce better Re-ID features in target domain than TAUDL. This is validated in Table~\ref{tab01}. Compared with TAUDL, our approach (BL+SOT+CAL-GRL$/$CAL-CCE) achieves considerable improvement in both tasks of ``Market1501$\rightarrow$Duke'' and ``Duke$\rightarrow$Market1501''. 
 
\begin{table}[htbp]
 \centering
 \caption{Comparison with the state-of-the-art approach of semi-supervised person Re-ID on Market-1501 and Duke.}
 \begin{adjustbox}{width=0.48\textwidth}
   \begin{tabular}{|c|cc|cc|}
   \toprule
   \multirow{2}[2]{*}{Method} & \multicolumn{2}{c|}{Duke$\rightarrow$Market1501} & \multicolumn{2}{c|}{Market1501$\rightarrow$Duke} \\
\cmidrule{2-5}         & mAP  & Rank-1 & mAP  & Rank-1 \\
   \midrule
   TAUDL~\cite{Li_2018_ECCV} & 41.2 & 63.7 & 43.5 & 61.7 \\
   \midrule
   BL+SOT+CAL-GRL & 46.6 & 72.2 & 44.3 & 62.0 \\
   BL+SOT+CAL-CCE (Ours) & \textcolor[rgb]{ 1, 0, 0}{\textbf{49.6}} & \textcolor[rgb]{ 1, 0, 0}{\textbf{73.7}} & \textcolor[rgb]{ 1, 0, 0}{\textbf{45.6}} & \textcolor[rgb]{ 1, 0, 0}{\textbf{64.0}} \\
   \bottomrule
   \end{tabular}%
   \end{adjustbox}
 \label{tab01}%
 \vspace*{-15pt}
\end{table}%
 
\begin{table}[htbp]
 \centering
 \caption{Comparison of camera-aware adversarial learning (CAL) and domain-aware adversarial learning (DAL) in unsupervised and semi-supervised settings on Market-1501 and Duke.}
 \begin{adjustbox}{width=0.48\textwidth}
   \begin{tabular}{|c|cc|cc|}
   \toprule
   \multirow{2}[2]{*}{Method} & \multicolumn{2}{c|}{Duke$\rightarrow$Market1501} & \multicolumn{2}{c|}{Market1501$\rightarrow$Duke} \\
\cmidrule{2-5}         & mAP  & Rank-1 & mAP  & Rank-1 \\
   \midrule
   BL+UOT+DAL & 25.5 & 54.1 & 26.2 & 42.4 \\
   \midrule
   BL+UOT+CAL-GRL & 30.9 & 60.4 & 31.0 & 47.7 \\
   BL+UOT+CAL-CCE (Ours) & \textcolor[rgb]{ 1, 0, 0}{\textbf{34.5}} & \textcolor[rgb]{ 1, 0, 0}{\textbf{64.3}} & \textcolor[rgb]{ 1, 0, 0}{\textbf{36.7}} & \textcolor[rgb]{ 1, 0, 0}{\textbf{55.4}} \\
   \midrule
   \midrule
  BL+SOT+DAL & 40.2 & 67.3 & 34.8 & 52.2 \\
  \midrule
  BL+SOT+CAL-GRL & 46.6 & 72.2 & 44.3 & 62.0 \\
  BL+SOT+CAL-CCE (Ours) & \textcolor[rgb]{ 1, 0, 0}{\textbf{49.6}} & \textcolor[rgb]{ 1, 0, 0}{\textbf{73.7}} & \textcolor[rgb]{ 1, 0, 0}{\textbf{45.6}} & \textcolor[rgb]{ 1, 0, 0}{\textbf{64.0}} \\
   \bottomrule
   \end{tabular}%
   \end{adjustbox}{}
 \label{tab03}%
 \vspace*{-10pt}
\end{table}%
 
To validate the effectiveness of the proposed camera-aware adversarial learning (CAL), we compare our approach with the state-of-the-art domain adaptation methods. They use domain-aware adversarial learning (DAL)~\cite{DBLP:conf/icml/GaninL15,DBLP:conf/cvpr/TzengHSD17} to conduct adversarial learning between source and target domains only. We implement DAL by ourselves to ensure a fair comparison. As in Table~\ref{tab03}, DAL (BL+UOT+DAL and BL+SOT+DAL) is inferior to the proposed CAL in both unsupervised and semi-supervised settings. There is a large gap between them in both experiments. This result further demonstrates the advantage of CAL by considering the discrepancy across all camera-level sub-domains.

\subsection{On the effectiveness of CAL and UOT}\label{sec:EXP-AS}
In the following ablation studies, we provide more details on the effectiveness of the two proposed components: camera-aware adversarial learning (CAL) and unsupervised online triplet generation (UOT) in Tables~\ref{tab02} and~\ref{tab05}.
 
\textbf{Effectiveness of CAL.}
First, BL+UOT+CAL-CCE consistently outperforms BL+UOT+CAL-GRL in Table~\ref{tab02}. The former improves $3.9\%$ ($64.3$ vs. $60.4$) and $7.7\%$ ($55.4$ vs. $47.7$) on Rank-1 accuracy in ``Duke$\rightarrow$Market1501'' and ``Market1501$\rightarrow$Duke''. This shows that the proposed CCE-based scheme can overcome the drawback of the GRL-based one, as analyzed in Section~\ref{sec:CAL}. Second, incorporating the proposed CAL (via either BL+UOT+CAL-CCE or BL+UOT+CAL-GRL in Table~\ref{tab02}) greatly improves over BL+UOT. This validates the effectiveness of CAL in helping reduce camera-level distribution discrepancy to learn better feature representation. Also, we validate the effectiveness of CAL in the semi-supervised setting. As seen in Table~\ref{tab05}, BL+SOT+CAL-CCE improves BL+SOT by $10.4\%$ ($49.6$ vs. $39.2$) and $6.4\%$ ($45.6$ vs. $39.2$) on mAP in ``Duke$\rightarrow$Market1501'' and ``Market1501$\rightarrow$Duke''. 
 
\textbf{Effectiveness of UOT.}
Adding the proposed unsupervised online triplets into the baseline (i.e., BL+UOT) clearly improves the baseline (BL), as seen in Table~\ref{tab02}. This confirms the benefit of exploiting discriminative information from target domain via the proposed UOT. In addition, we test two schemes to compute the distance matrix ${\mathbf M}$ in Section~\ref{sec:UOT}, via Euclidean distance (BL+UOT(eud)) and the default re-ranking algorithm (BL+UOT), respectively. As seen, BL+UOT does achieve better performance.

In addition, we are interested in what if CAL is used alone without the UOT component. Both BL+CAL-GRL and BL+CAL-CCE are investigated at the bottom of Table~\ref{tab02}. As seen, merely using CAL is insufficient. They do not show sufficient improvement over BL+UOT or even BL+UOT(eud). BL+CAL-CCE even fails in the case of ``Duke$\rightarrow$Market1501'', although showing some improvement on ``Market1501$\rightarrow$Duke''. This result is explained as follows. Duke has more cameras than Market1501. Also, in the literature, when the same Re-ID model is applied to them, the performance on Duke is usually inferior to that on Market1501~\cite{DBLP:conf/cvpr/LiZXW14,DBLP:conf/cvpr/AhmedJM15,DBLP:conf/iccv/ZhengZY17,kalayeh2018human}. This shows that Market1501 is a relatively easier dataset than Duke. When Market1501 is target domain and the more challenging Duke is source domain, its distribution could be altered significantly in order to fit the Duke's, if no discriminative information from target domain is used as a regularizer. Meanwhile, the situation will become less critical in the other way round. This is because Duke's data distribution is more complicated and therefore has sufficient ``capacity'' to fit Market1501's. 
 
The above result clearly shows the necessity of UOT in our framework. Working together, the two proposed components produce the best performance in Table~\ref{tab02}.

\begin{table}[htbp]
 \centering
 \caption{Performance of the proposed framework when using different components on Market1501 and DukeMTMC-reID (Duke).}
 \begin{adjustbox}{width=0.48\textwidth}
   \begin{tabular}{|c|cc|cc|}
   \toprule
   \multirow{2}[1]{*}{Method} & \multicolumn{2}{c|}{Duke$\rightarrow$Market1501} & \multicolumn{2}{c|}{Market1501$\rightarrow$Duke} \\
\cmidrule{2-5}         & mAP  & Rank-1 & mAP  & Rank-1 \\
   \midrule
   BL & 19.4 & 47.1 & 21.3 & 38.4 \\
\midrule
   BL+UOT(eud) & 23.6 & 51.0 & 24.1 & 40.2 \\
   BL+UOT & 27.4 & 55.5 & 27.5 & 44.3 \\
   \midrule
   BL+UOT+CAL-GRL & 30.9 & 60.4 & 31.0 & 47.7 \\
   BL+UOT+CAL-CCE (Ours) & \textcolor[rgb]{ 1, 0, 0}{\textbf{34.5}} & \textcolor[rgb]{ 1, 0, 0}{\textbf{64.3}} & \textcolor[rgb]{ 1, 0, 0}{\textbf{36.7}} & \textcolor[rgb]{ 1, 0, 0}{\textbf{55.4}} \\
   \midrule
   \midrule
   BL+CAL-GRL & 20.5 & 47.6 & 22.7 & 41.4 \\
   BL+CAL-CCE & 8.4 & 27.6 & 23.8 & 45.4 \\
   \bottomrule
   \end{tabular}%
    \end{adjustbox}
      \label{tab02}%
      \vspace*{-15pt}
\end{table}%
 
\begin{table}[htbp]
 \centering
 \caption{Effectiveness of CAL in the semi-supervised setting on Market-1501 and DukeMTMC-reID (Duke).}
 \begin{adjustbox}{width=0.48\textwidth}
   \begin{tabular}{|c|cc|cc|}
   \toprule
   \multirow{2}[2]{*}{Method} & \multicolumn{2}{c|}{Duke$\rightarrow$Market1501} & \multicolumn{2}{c|}{Market1501$\rightarrow$Duke} \\
\cmidrule{2-5}         & mAP  & Rank-1 & mAP  & Rank-1 \\
   \midrule
    BL+SOT & 39.2 & 65.9 & 39.2 & 56.6 \\
    \midrule
   BL+SOT+CAL-GRL & 46.6 & 72.2 & 44.3 & 62.0 \\
   BL+SOT+CAL-CCE (Ours)& \textcolor[rgb]{ 1, 0, 0}{\textbf{49.6}} & \textcolor[rgb]{ 1, 0, 0}{\textbf{73.7}} & \textcolor[rgb]{ 1, 0, 0}{\textbf{45.6}} & \textcolor[rgb]{ 1, 0, 0}{\textbf{64.0}} \\
   \bottomrule
   \end{tabular}%
   \end{adjustbox}
 \label{tab05}%
 \vspace*{-10pt}
\end{table}%
 
\subsection{Further evaluation of the proposed framework}\label{sec:EXP-FECAL}
 
\begin{table}[htbp]
 \centering
 \caption{The data distribution discrepancy of inter-domain (between source and target domains) and inter-camera (across all cameras in target domain) on the task of Duke$\rightarrow$Market1501. Note that a smaller value indicates better performance in this table.}
 \begin{adjustbox}{width=0.45\textwidth}
   \begin{tabular}{|l|c|c|c|c|}
   \toprule
         & BL   & DAL  & CAL-GRL & CAL-CCE \\
   \midrule
   Inter-domain ($\times 10^3$) & 1.48 & 1.22 & 1.28 & 1.25 \\
   \midrule
   Inter-camera ($\times 10^2$) & 6.76 & 5.97 & 4.36 & 2.90 \\
 \midrule
   \end{tabular}%
    \end{adjustbox}
 \label{tab06}%
 \vspace*{-15pt}
\end{table}%

\begin{figure}
\centering
\subfigure[Baseline (BL)]{
\includegraphics[width=3.5cm]{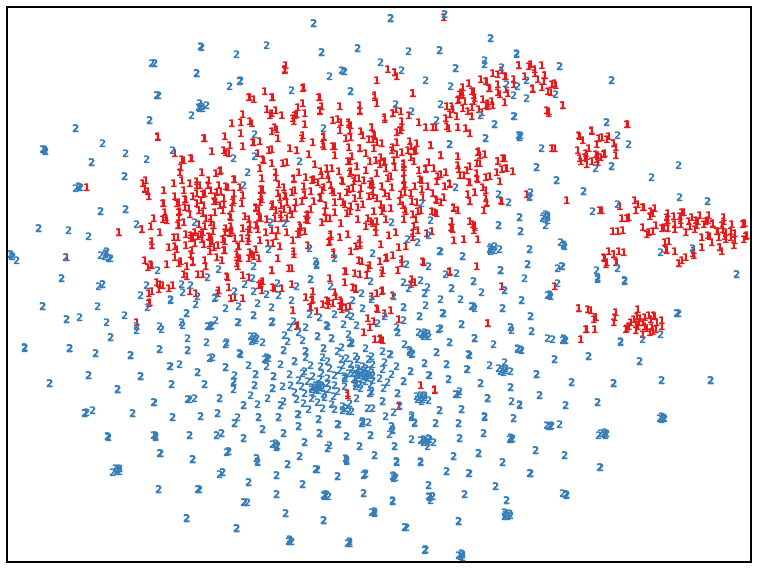}
}
\subfigure[CAL-CCE (Ours)]{
\includegraphics[width=3.5cm]{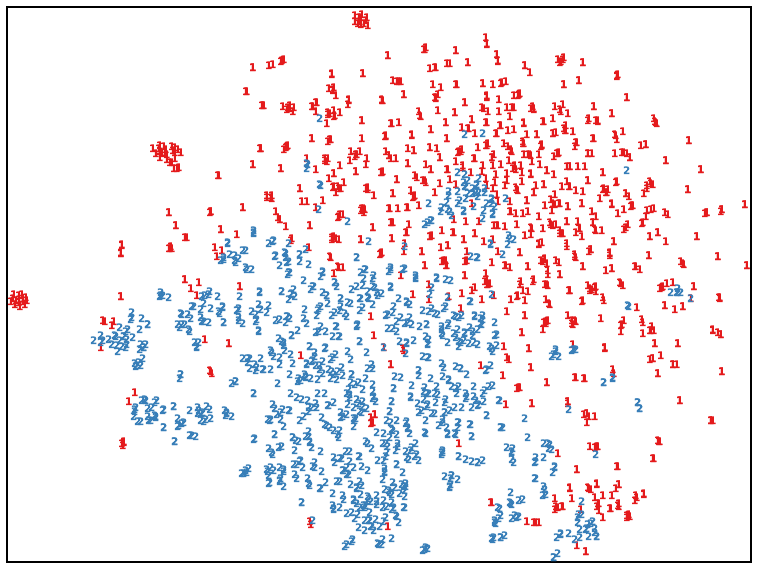}
}
\subfigure[Baseline (BL)]{
\includegraphics[width=3.5cm]{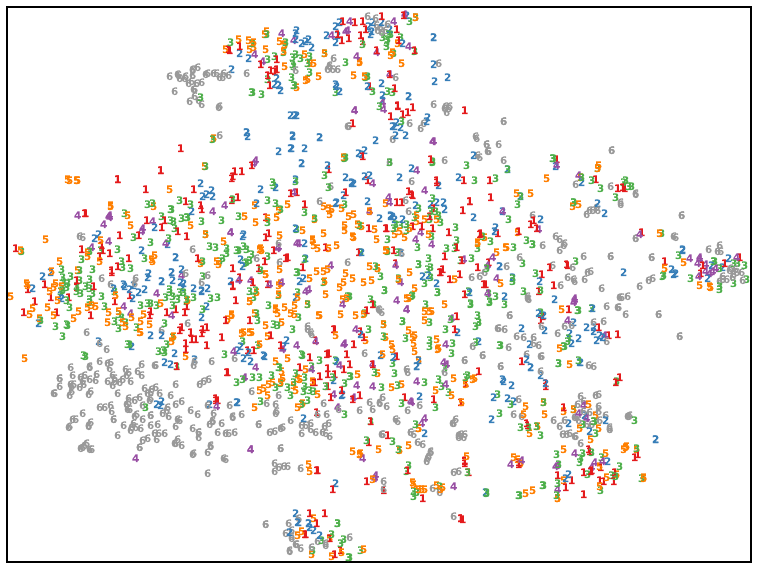}
}
\subfigure[CAL-CCE (Ours)]{
\includegraphics[width=3.5cm]{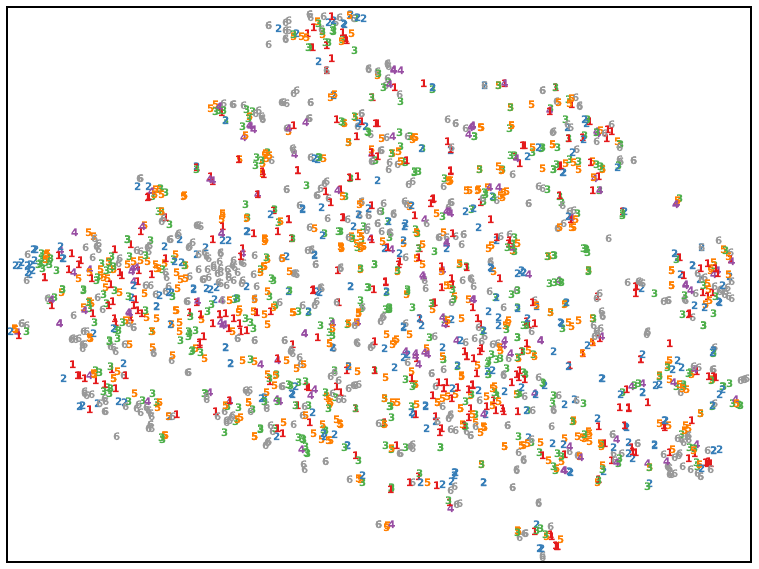}
}
\caption{Visualization of data distributions at the domain-level and camera-level via t-SNE~\cite{maaten2008visualizing}. The features of each image are extracted by the baseline (BL) and CAL-CCE in the task of ``DukeMTMC-reID$\rightarrow$Market1501'', respectively. The top shows the distributions of source and target domains (i.e., inter-domain).
 The bottom illustrates the distribution of each camera class in target domain (i.e., inter-camera on Market1501), where different colors denote different camera classes. As seen, the proposed CAL-CCE effectively ``mixes'' the two domains and the camera classes as expected. Illustration of the methods of DAL and CAL-GRL is in the supplement material.}
\label{fig9}
\vspace*{-15pt}%
\end{figure}

We examine the inter-domain (between source and target domains) and inter-camera (across all cameras in target domain) discrepancy to validate the effectiveness of CAL, as reported in Table~\ref{tab06}. Since our goal is to obtain better feature representation for target domain, we focus on target domain when examining the discrepancy across all cameras. In this experiment, we measure the inter-domain discrepancy by the distance $d_{\mathrm{inter-domain}}=\left \| \overline{X}_{s}-\overline{X}_{t} \right \|_{2}$, where $\overline{X}_{s}$ and $\overline{X}_{t}$ denote the sample mean of source and target domains, respectively. To measure the inter-camera discrepancy, we use $d_{\mathrm{inter-camera}}=\frac{1}{C_{t}}\sum_{c=1}^{C_{t}}\left \| \overline{X}_{t,c}-\overline{X}_{t} \right \|_{2}$, where $\overline{X}_{t,c}$ is the sample mean of the $c$th camera class in target domain and $C_{t}$ is the total number of cameras in target domain. These distances are calculated in Table~\ref{tab06}. First, for the inter-domain discrepancy, DAL, CAL-GRL and CAL-CCE are all smaller than BL. This validates that they are all able to reduce the discrepancy between source and target domains. Particularly, since DAL specifically focuses on the overall domain-level discrepancy, its distance is smaller than those of CAL-GRL and CAL-CCE. 
In addition, CAL-GRL has a slightly larger value than CAL-CCE. This is consistent with the analysis on the drawback of CAL-GRL in Section~\ref{sec:CAL}. Second, for the inter-camera discrepancy, both CAL-GRL and CAL-CCE achieve smaller distances than DAL because DAL does not consider camera-level discrepancy. 
Besides, this experiment shows that CAL-CCE achieves the smallest distance, showing its best capability in reducing the discrepancy across the cameras in target domain. Additionally, we visualize the data distributions obtained by the feature representation from BL and CAL-CCE in Fig.~\ref{fig9}. The result further illustrates the effectiveness of CAL-CCE. 

\begin{table}[htbp]
 \centering
 \caption{Comparison of the CCE-based scheme and the AOE-based scheme on Market1501 and DukeMTMC-reID (Duke).}
   \begin{adjustbox}{width=0.4\textwidth}
   \begin{tabular}{|c|cc|cc|}
   \toprule
   \multirow{2}[1]{*}{Method} & \multicolumn{2}{c|}{Duke$\rightarrow$Market1501} & \multicolumn{2}{c|}{Market1501$\rightarrow$Duke} \\
\cmidrule{2-5}         & mAP  & Rank-1 & mAP  & Rank-1 \\
   \midrule
   AOE   & 29.6 & 59.9 & 31.5 & 51.3 \\
   \midrule
   CCE (Ours) & \textcolor[rgb]{ 1, 0, 0}{\textbf{34.5}} & \textcolor[rgb]{ 1, 0, 0}{\textbf{64.3}} & \textcolor[rgb]{ 1, 0, 0}{\textbf{36.7}} & \textcolor[rgb]{ 1, 0, 0}{\textbf{55.4}} \\
   \bottomrule
   \end{tabular}%
   \end{adjustbox}
 \label{tab10}%
 \vspace*{-10pt}
\end{table}%

As analyzed in the \textit{Remarks} of Section~\ref{sec:CAL}, why do we prefer the CCE-based scheme rather than equiprobably mixing a camera class with all the other ($C_{s} + C_{t}-1$) ones? In this experiment, we name this setting AOE, standing for ``all others equiprobability.'' We compare AOE and CCE in Table~\ref{tab10}. As shown, the CCE-based scheme outperforms the AOE-based scheme in both tasks. This is consistent with our previous analysis. It indicates that during reducing camera-level discrepancy, giving the reduction of cross-domain discrepancy higher priority is beneficial because it is usually more significant than the within-domain counterpart and affects the performance more.

Finally, we validate the advantage of the unsupervised \textit{online} triplet (UOT) generation by comparing it with the \textit{offline} way. To generate offline triplets, we use the same method in Section~\ref{sec:UOT} with the mere difference that they are generated by the feature from the baseline model before training starts.
In total, 88,719 and 100,742 triplets are generated on Market1501 and Duke. We randomly select $40$ triplets in each batch (i.e., $120$ samples, giving the same batch size as the online method) to train our model. As seen in Table~\ref{tab09}, offline method is clearly inferior. It confirms the advantage of online method by utilizing the steadily improved features in training, as discussed in Section~\ref{sec:UOT}.
 
\begin{table}[htbp]
 \centering 
 \caption{Comparison of offline and online triplet generation on Market-1501 and DukeMTMC-reID (Duke).}
  \begin{adjustbox}{width=0.48\textwidth}
   \begin{tabular}{|c|cc|cc|}
   \toprule
   \multirow{2}[1]{*}{Method} & \multicolumn{2}{c|}{Duke$\rightarrow$Market1501} & \multicolumn{2}{c|}{Market1501$\rightarrow$Duke} \\
\cmidrule{2-5}         & mAP  & Rank-1 & mAP  & Rank-1 \\
     \midrule
   Offline Triplets & 13.5 & 37.1 & 10.2 & 21.1 \\
   \midrule
   Online Triplets (Ours) & \textcolor[rgb]{ 1, 0, 0}{\textbf{27.4}} & \textcolor[rgb]{ 1, 0, 0}{\textbf{55.5}} & \textcolor[rgb]{ 1, 0, 0}{\textbf{27.5}} & \textcolor[rgb]{ 1, 0, 0}{\textbf{44.3}} \\
   \bottomrule
   \end{tabular}%
   \end{adjustbox}
 \label{tab09}%
 \vspace*{-20pt}
\end{table}%
 
\section{Conclusion}
This paper proposes a novel deep domain adaptation framework to address two key issues in unsupervised cross-domain person Re-ID. It clearly shows that when pursuing better feature representation for person Re-ID, considering camera-level domain discrepancy is beneficial. Also, exploring discriminative information from unlabeled target domain is equally, if not more, important. Only when these two components are adequately resolved, unsupervised cross-domain person Re-ID can become promising. In the future work, we will further develop more flexible and adaptive camera-aware domain adaptation for this task.
 
{\small
\bibliographystyle{ieee_fullname}
\bibliography{egbib}
}

\textbf{\LARGE{Supplementary material}}

\setcounter{equation}{7}
\renewcommand\thesection{A1}
\section{Theoretical analysis of the CCE-based scheme}

The following conducts theoretical analysis for the proposed CCE- (i.e., cross-domain camera equiprobability) based scheme. 


\textbf{Proposition.} \textit{Let ${\mathcal S}$ and ${\mathcal T}$ denote source and target domains. $x^s$ and $x^t$ are the images from the two domains; $p_s(x)$ and $p_t(x)$ are their probability density functions; and $C_s$ and $C_t$ are the number of camera classes in these two domains. Let $p(x|\mathcal{C}^s_i)$ and $p(x|\mathcal{C}^t_i)$ be the class-conditional density functions of the $i$th camera class in source and target domains, respectively. It can be proved that ideally, minimizing the CCE loss will lead to}
\begin{eqnarray}\label{eqn:proposition} 
p(x^s|\mathcal{C}^t_i) &=& p_s(x^s),\quad \forall x^s \in {\mathcal S};~i=1,\cdots,C_t.\\~\nonumber
p(x^t|\mathcal{C}^s_i) &=& p_t(x^t),\quad \forall x^t \in {\mathcal T};~i=1,\cdots,C_s.\\~\nonumber
p_s(x) &=& p_t(x), \quad \forall x \in {\mathcal S}\cup{\mathcal T}.
\end{eqnarray}

\textbf{Proof.} All the following analysis is conducted in the context of the learned feature representation (or equally, the learned shared subspace). Given an image $x^s$ from source domain, its posteriori probability with respect to the $i$th camera class in target domain (denoted by $\mathcal{C}^t_i (i=1,\cdots,C_t)$) can be expressed via the Bayes' rule as 
\begin{equation}\label{eqn:Bayes}
P(\mathcal{C}^t_i|x^s) = \frac{p(x^s|\mathcal{C}^t_i)P(\mathcal{C}^t_i)}{p_s(x^s)},\quad \forall x^s \in {\mathcal S};~i=1,\cdots,C_t, 
\end{equation}where $p(x^s|\mathcal{C}^t_i)$ is the class-conditional probability density function of the $i$th camera class in target domain, $p_s(x^s)$ denotes the probability density function of the images in source domain, and $P(\mathcal{C}^t_i)$ is the priori probability of the $i$th camera class in target domain. Referring to Eq.(3) in our paper, $P(\mathcal{C}^t_i|x^s)$ is just $D(B(x),j)$ in the CCE loss. 

Let us investigate the CCE loss in Eq.(3) of our paper to gain understanding on the optimal value of $D(B(x),j)$ when this loss is minimized. Since the CCE loss is defined for each individual image $x$ independently, it will be sufficient to investigate the minimization of the loss for any given image $x$. Without loss of generality, it is assumed that $x$ is from source domain. For clarity, $D(B(x),j)$ is compactly denoted by $D_j$. With respect to $D_j (j=1,\cdots,C_t)$, the minimization of the CCE loss can be expressed as a constrained optimization 
\begin{equation}\label{eqn:opt}
\underset{\{D_1,\cdots,D_{C_t}\}}\min\left( -\frac{1}{C_{t}}\sum_{j=1}^{C_{t}}\log (D_j)  \right)
\end{equation} with the constraints of $D_j\geq{0}$ and $\sum_{j=1}^{C_{t}}D_j=1$, considering that $D_j$ represents the posteriori probability. Due to the symmetry of the objective function with respect to the variables $D_1,\cdots,D_{C_t}$, it is not difficult to see that the optimal value of $D_j$ is $1/C_t$ for $j=1,\cdots,C_t$. A rigorous proof can be readily obtained by applying the Karush-Kuhn-Tucker conditions to this optimization, which is omitted here. This indicates that $P(\mathcal{C}^t_i|x^s)$ will equal $1/C_t$ when the CCE loss is minimized for this given image $x$. Now we assume the ideal case that this CCE loss is minimized for any given image $x$ in source domain\footnote{Note that such an ideal case may not be really achieved in practice. Nevertheless, it helps to clearly reveal the effect of minimizing the CCE loss in the theoretical sense.}.   

Let us turn to Eq.(\ref{eqn:Bayes}) and rearrange it as
\begin{equation}\label{eqn:Bayes-1}
p(x^s|\mathcal{C}^t_i) = \frac{P(\mathcal{C}^t_i|x^s)}{P(\mathcal{C}^t_i)}p_s(x^s),\quad \forall x^s \in {\mathcal S};~i=1,\cdots,C_t. 
\end{equation}Without loss of generality, equal priori probability can be set for the $C_t$ camera classes in target domain, that is, $P(\mathcal{C}^t_i)$ is constant $1/C_t$. Further, note that by optimizing $D_j$ in Eq.(\ref{eqn:opt}) above, it can be known that 
\begin{equation}\label{eqn:Bayes-1-1}
P(\mathcal{C}^t_i|x^s)=\frac{1}{C_t},\quad \forall x^s \in {\mathcal S};~i=1,\cdots,C_t.  
\end{equation}
Combining the above results, Eq.(\ref{eqn:Bayes-1}) becomes
\begin{equation}\label{eqn:Bayes-2}
p(x^s|\mathcal{C}^t_i) = \frac{1/C_t}{1/C_t}p_s(x^s)=p_s(x^s),\quad \forall x^s \in {\mathcal S};~i=1,\cdots,C_t. 
\end{equation}Note that the right hand side of this equation does not depend on the index $i$. This indicates that \textit{in the learned shared subspace, the class-conditional density function for each camera class in \underline{target} domain becomes same for any given $x^s \in {\mathcal S}$}. 
Applying the same argument to the images in target domain can similarly obtain 
\begin{equation}\label{eqn:Bayes-3}
p(x^t|\mathcal{C}^s_i) = p_t(x^t),\quad \forall x^t \in {\mathcal T};~i=1,\cdots,C_s, 
\end{equation}where $p(x^t|\mathcal{C}^s_i)$ and $p_t(x^t)$ are defined in the similar way as the above. This result indicates that \textit{in the learned shared subspace, the class-conditional density function for each camera class in \underline{source} domain becomes same for any given $x^t \in {\mathcal T}$}. 

The above results indicate that for an image in source domain, it will not feel the distribution discrepancy among the camera classes in target domain. Furthermore, its class-conditional density function value (e.g., $p(x^s|\mathcal{C}^t_i)$) for those camera classes just equals its density function value in source domain (e.g., $p_s(x^s)$). The similar remark can be made for an image in target domain. 

Upon the above results, the following further proves that in the learned shared subspace, the data distributions of source and target domains will become identical and this removes the domain-level distribution discrepancy. 

For any image $x^s$ from source domain, its value evaluated by the probability density function of target domain can be obtained as
\begin{equation}\label{eqn:Bayes-4}
p_t(x^s) = \sum_{i=1}^{C_t}p(x^s|C_{i}^t)P(C_i^t) = \sum_{i=1}^{C_t}p_s(x^s)P(C_i^t) = p_s(x^s), 
\end{equation}where the first equality is due to Eq.(\ref{eqn:Bayes-2}) and the second one is because $\sum_{i=1}^{C_t}P(C_i^t)=1$. Similarly, the result for any given image $x^t$ from target domain can be obtained as
\begin{equation}\label{eqn:Bayes-5}
p_s(x^t) = \sum_{i=1}^{C_s}p(x^t|C_{i}^s)P(C_i^s) = \sum_{i=1}^{C_s}p_t(x^t)P(C_i^s) = p_t(x^t), 
\end{equation}where the first equality is due to Eq.(\ref{eqn:Bayes-3}) and the second one is because $\sum_{i=1}^{C_s}P(C_i^s)=1$. 

Collectively, the above two results indicate that for any image $x$ from either source or target domain, the following result can be obtained. 
\begin{equation}\label{eqn:Bayes-5}
p_s(x) = p_t(x), \quad \forall x \in {\mathcal S}\cup{\mathcal T}.
\end{equation} 
This means that the two distributions, $p_s(x)$ and $ p_t(x)$, are identical on the set ${\mathcal S}\cup{\mathcal T}$. With respect to the definitions of the two distributions, this indicates that \textit{upon the learned feature representation, the data distributions of source and target domains become identical on the set ${\mathcal S}\cup{\mathcal T}$ and that the distribution discrepancy is therefore removed}. \quad\quad\quad$\blacksquare$

In addition, it is worth mentioning that the ideal minimization of the CCE loss does not theoretically guarantee that in the shared subspace, an image from either source or target domain will not feel the distribution discrepancy among the camera classes in \textit{its own} domain. In other words, the results that $p(x^s|\mathcal{C}^s_1)=\cdots=p(x^s|\mathcal{C}^s_{C_s})$ or $p(x^t|\mathcal{C}^t_1)=\cdots=p(x^t|\mathcal{C}^t_{C_t})$ cannot directly be derived from the ideal minimization of the CCE loss. 

Nevertheless, note that Eq.(\ref{eqn:Bayes-5}) implies that at any place $x$ in ${\mathcal S}\cup{\mathcal T}$, the probability density of images from source domain is the same as the probability density of images form target domain. This means that the images from two domains have been adequately mixed up. In this case, considering that the result $p(x^s|\mathcal{C}^t_1)=\cdots=p(x^s|\mathcal{C}^t_{C_t})$ is true (as proved in Eq.(\ref{eqn:Bayes-2})), we can reasonably expect that this result shall be generalized from $x^s$ to $x^t$, that is, $p(x^t|\mathcal{C}^t_1)=\cdots=p(x^t|\mathcal{C}^t_{C_t})$ becomes true. Applying the same argument can obtain the result $p(x^s|\mathcal{C}^s_1)=\cdots=p(x^s|\mathcal{C}^s_{C_s})$. Therefore, it can be reasonably expected that in practice in the shared subspace, an image from either source or target domain will not feel the distribution discrepancy among the camera classes in \textit{its own} domain. Experimental study has been conducted to show that this tendency can indeed be observed in practice, as shown by Table 6 in our paper and the following Fig.~\ref{fig9}.

\begin{figure*}[htb]
\renewcommand\thefigure{A1}
\centering
\subfigure[Baseline]{
\includegraphics[width=3.5cm]{fig/fig9-1.pdf}
}
\subfigure[DAL]{
\includegraphics[width=3.5cm]{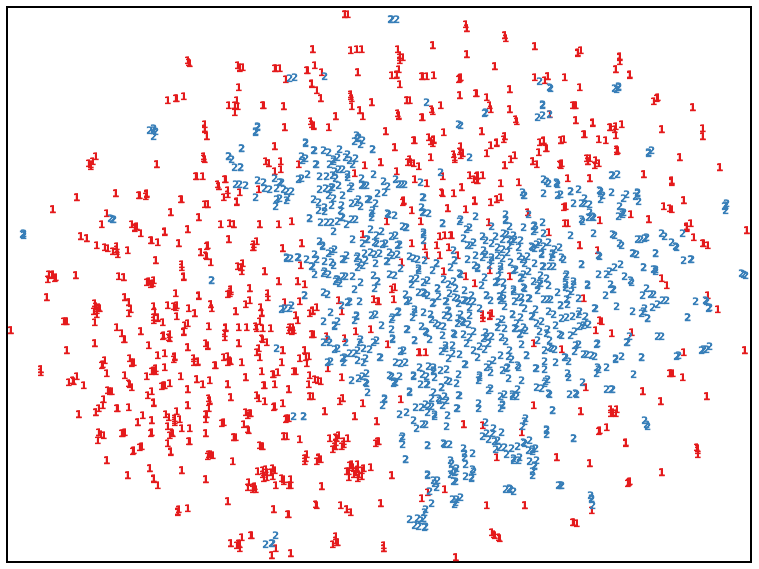}
}
\subfigure[CAL-GRL]{
\includegraphics[width=3.5cm]{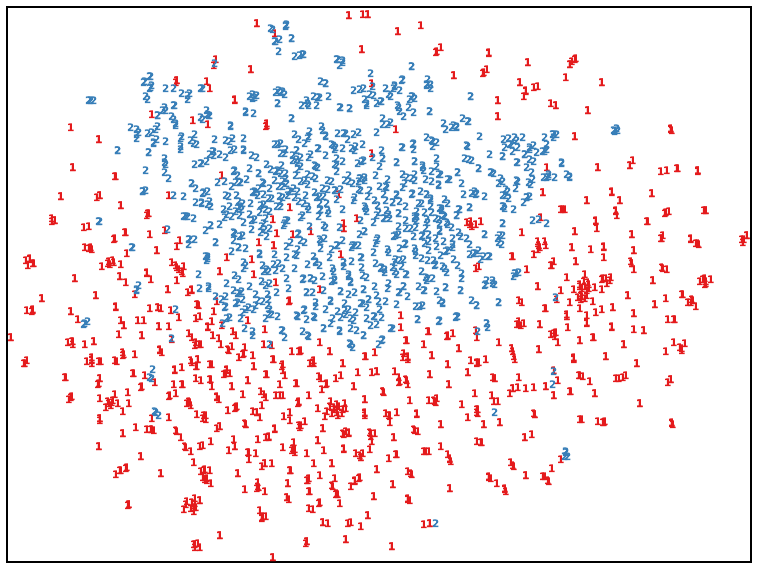}
}
\subfigure[CAL-CCE]{
\includegraphics[width=3.5cm]{fig/fig9-4.pdf}
}
\subfigure[Baseline]{
\includegraphics[width=3.5cm]{fig/fig9-5.pdf}
}
\subfigure[DAL]{
\includegraphics[width=3.5cm]{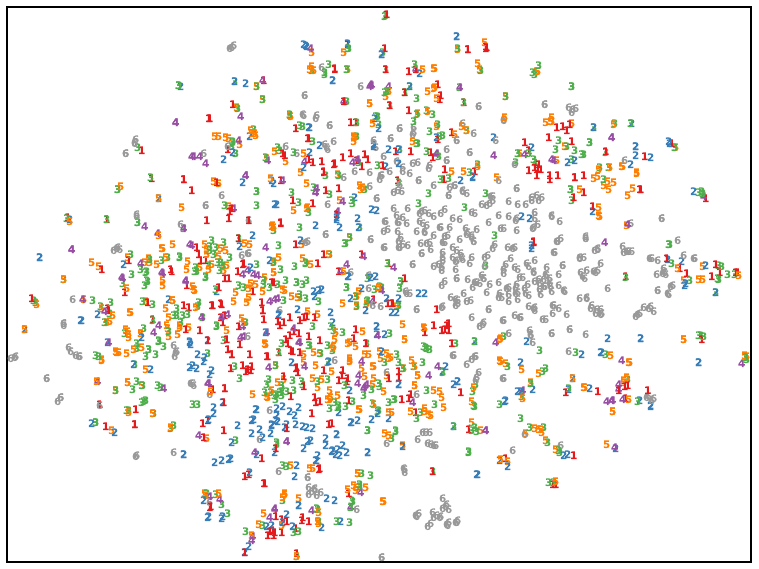}
}
\subfigure[CAL-GRL]{
\includegraphics[width=3.5cm]{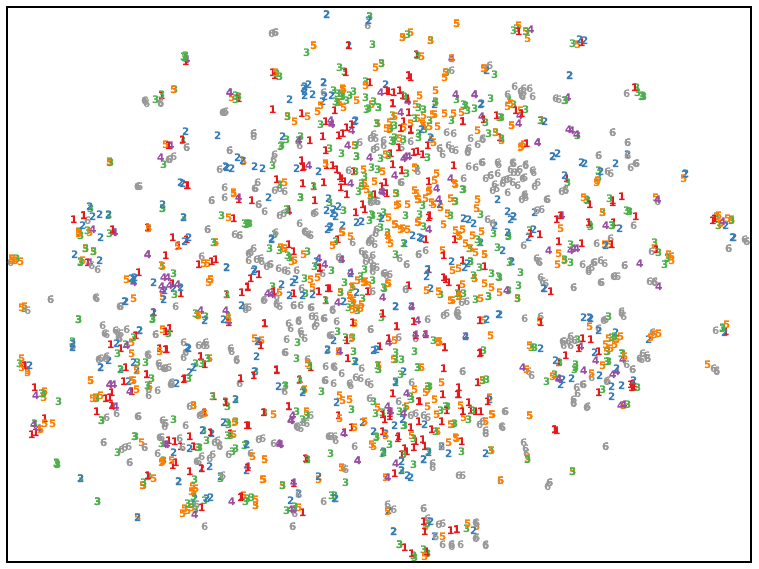}
}
\subfigure[CAL-CCE]{
\includegraphics[width=3.5cm]{fig/fig9-8.pdf}
}
\caption{Visualization of data distributions at the domain-level and camera-level via t-SNE~\cite{maaten2008visualizing}. The features of each image are extracted by the baseline (BL), DAL, CAL-GRL and CAL-CCE in the task of ``DukeMTMC-reID$\rightarrow$Market1501'', respectively.  The top shows the distributions of source and target domains (i.e., inter-domain), where blue and red colors indicate source and target domains, respectively. The bottom illustrates the distribution of each camera class in target domain (i.e., inter-camera on Market1501), where different colors denote different camera classes. Note that all figures correspond to the experimental results in Table 6 of our paper.}
\label{fig19}
\end{figure*}

\renewcommand\thesection{A2}
\section{Visualization of data distributions}
The data distributions are visualized at the domain-level and camera-level via t-SNE~\cite{maaten2008visualizing} in Fig.~\ref{fig19}. We extract the features of each image by the baseline model (BL), DAL, CAL-GRL and CAL-CCE, respectively, in the task of ``DukeMTMC-reID$\rightarrow$Market1501''.  The top row shows the distributions of source and target domains (i.e., inter-domain), where blue and red colors indicate source and target domains, respectively. The bottom row illustrates the distribution of each camera class in target domain (i.e., inter-camera on Market1501), where different colors denote different camera classes. 

First, from the inter-domain results shown in the top row of Fig.~\ref{fig19}, it can be seen that DAL, CAL-GRL and CAL-CCE can effectively ``mix'' the two domains when compared with BL. This validates that they are all able to reduce the data distribution discrepancy between source and target domains. 

Second, from the inter-camera result shown in the bottom row in Fig.~\ref{fig19}, it can be seen that both CAL-GRL and CAL-CCE seem to better ``mix'' these camera classes than BL and DAL which do not consider any camera-level discrepancy. 
Furthermore, consistent with its lowest inter-camera distance reported in Table 6 of our paper, CAL-CCE displays an excellent ``mixture'' of different camera classes as expected, further illustrating its best capability in reducing the camera-level discrepancy in target domain. 
\begin{figure}[htb]
\renewcommand\thefigure{A2}
\centering
\subfigure[$k$~(It decides $k_p$)]{
\includegraphics[width=3.8cm]{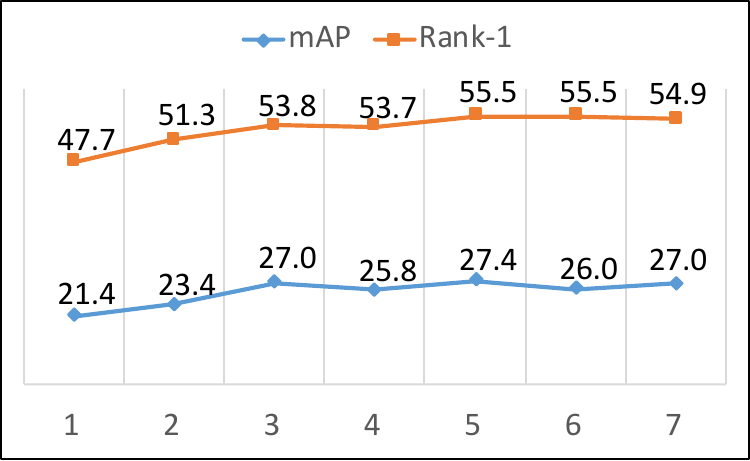}
}
\subfigure[$k_n$]{
\includegraphics[width=3.8cm]{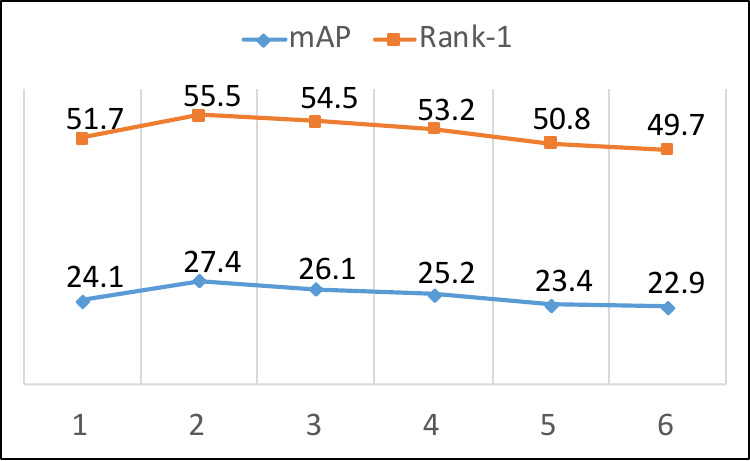}
}
\caption{\footnotesize Parameter sensitivity (by \underline{BL+UOT} in ``Duke$\rightarrow$Market1501'').}
\label{fig10}
\end{figure}

\renewcommand\thesection{A3}
\section{On the parameter sensitivity of UOT.}
In this section, we conduct experiments to observe the parameter sensitivity of UOT. The results are reported in Fig.~\ref{fig10}. Firstly, note that $k_p$ is not directly preset but decided by $k$ and the specific data (Recall that $k_p$ is the number of positive samples of an anchor within the top-$k$ positions and share its fragment). As in Fig.~\ref{fig10}, when $k$ is as small as 1, no positive samples (i.e., $k_p$ is often zero) could be found and thus performance is poor. When $k$ goes up to $5$, the performance tends to plateau. Secondly, for $k_n$, if we only select one negative sample (i.e., $k_n$ is set as 1), this sample often undesirably shares the same identity as the anchor. Meanwhile, setting $k_n$ too large could include many easy negative samples, instead of hard negative ones preferred by UOT. In all experiments, we uniformly set $k$ and $k_n$ as $5$ and $2$.

\end{document}